\documentclass[letterpaper, 10 pt, conference]{ieeeconf}    % Comment this line out if you need a4paper

\IEEEoverridecommandlockouts                              % This command is only needed if 
\usepackage{graphicx}
\usepackage{lineno}
\usepackage[hidelinks]{hyperref}
\usepackage{xcolor}
\usepackage{pgfplots}
\pgfplotsset{compat=1.18} 
\usepackage{tikz}
\usetikzlibrary{positioning}
\usetikzlibrary{shapes.geometric}
\usetikzlibrary{shapes.misc}
\usetikzlibrary{external}
\usepackage{paralist}
\usepackage{mathtools}

\usepackage{xcolor}
\usepackage{siunitx}
\usepackage{multirow}
\usepackage{epstopdf}
\usepackage{todonotes}
\usepackage{epstopdf}
\usepackage{epsfig}
\usepackage{float}
\usepackage{amsmath} % assumes amsmath package installed
\usepackage{amssymb}  % assumes amsmath package installed
\usepackage{mathtools}
\usepackage[acronym,nomain,nonumberlist]{glossaries}
\usepackage{url}
\usepackage{algorithm}
 \usepackage{algpseudocode}

 \newlength\fwidth

 \usepackage{pgfplots}
\pgfplotsset{compat=newest}
\usepackage{tikz}
\usetikzlibrary{positioning}
\usetikzlibrary{shapes.geometric}
\usetikzlibrary{shapes.misc}
\usetikzlibrary{external}
\usetikzlibrary{shapes,arrows,patterns}
\usetikzlibrary{arrows.meta}
\usetikzlibrary{plotmarks}
\usetikzlibrary{calc,decorations.markings,math}

\usepackage{placeins} 
\usepackage{tabularx}

\usepackage{multirow}\usepackage{adjustbox} 
 \usepackage{array} 
 \newcolumntype{R}[2]{% 
 >{\adjustbox{angle=#1,lap=\width-(#2)}\bgroup}% 
   c%     
 <{\egroup}% 
 } 
 
 \newcommand*\rot{\multicolumn{1}{R{90}{1em}}}% no optional argument here, please! 
\graphicspath{ {pictures/} }
\newacronym{lidar}{LiDAR}{Light Detection and Ranging}
\newacronym{radar}{RADAR}{Radio Detection and Ranging}
\newacronym{av2}{AV2}{Argoverse2}
\newacronym{sota}{SOTA}{state-of-the-art}
\newacronym{bev}{BEV}{bird's eye view}
\newacronym{dnn}{DNN}{Deep Neural Network}
\newacronym{mvp}{MVP}{Multimodal Virtual Points}
\newacronym{map}{mAP}{mean Average Precision}
\newacronym{ap}{AP}{Average Precision}
\newacronym{cds}{CDS}{Composite Detection Score}
\newacronym{rpn}{RPN}{Region Proposal Networks}
\newacronym{fsd}{FSD}{Fully Sparse Detector}
\newacronym{fsf}{FSF}{Fully Sparse Fusion}

\newacronym{htc}{HTC}{Hybrid Task Cascade}
\newacronym{cbgs}{CBGS}{Class-balanced grouping and sampling}
\newacronym{rbgs}{RBGS}{Range-balanced grouping and sampling}

\newacronym{sir}{SIR}{Sparse Instance Recognition}
\newacronym{sph}{SPH}{sparse prediction head}
\newacronym{mlp}{MLP}{Multi Layer Perceptrons}
\newacronym{soa}{SoA}{State of the Art}
\newacronym{ccl}{CCL}{Connected Component Labelling}
\newacronym{vvm}{VVM}{Virtual Voxel Mixer}
\newacronym{re}{RE}{Range Expert}

\newacronym{nds}{NDS}{NuScenes Detection Score}

\usepackage{amsmath} % assumes amsmath package installed
\usepackage[nameinlink,capitalise]{cleveref}
\usepackage{subcaption} % For subfigures and subcaptions

\title{\LARGE \bf
Towards Long-Range 3D Object Detection for 
Autonomous
Vehicles
}

\author{Ajinkya Khoche$^{1,2}$, Laura Pereira Sánchez$^{3}$, Nazre Batool$^{2}$, Sina Sharif Mansouri$^{2}$ and Patric Jensfelt$^{1}$% <-this % stops a space
\thanks{*This work was supported by the research grant PROSENSE (2020-02963) funded by VINNOVA. %and by the Wallenberg AI, Autonomous Systems and Software Program (WASP) funded by the Knut and Alice Wallenberg Foundation
}% <-this % stops a space
\thanks{$^{1}$KTH Royal Institute of Technology, Stockholm 10044, Sweden. Corresponding author's e-mail: {\tt\small khoche@kth.se}}%
\thanks{$^{2}$Autonomous Transport Solutions Lab, Scania Group, Södertälje, SE-15139, Sweden}%The author is employed at Scania CV AB, 151 87 Sodertälje, Sweden.}%
\thanks{$^{3}$Stockholm University, Fysikum, Stockholm 114 21, Sweden.}
}

\begin{document}

%%%%%%%%%%%%%%%%%%%%%%%%%%%%%

\maketitle
\thispagestyle{empty}
\thispagestyle{empty}
\pagestyle{empty}

%%%%%%%%%%%%%%%%%%%%%%%%%%%%%
\begin{abstract}
% We propose an innovative solution to address the crucial need for long-range perception in autonomous vehicles, particularly on highways. 
%Long-range perception plays a pivotal role in 
3D object detection at long-range is crucial for ensuring the safety and efficiency of self-driving vehicles, allowing them to accurately perceive and react to objects, obstacles, and potential hazards from a distance. 
But most current state-of-the-art LiDAR based methods are range limited due to sparsity at long-range, which generates a form of domain gap between points closer to and farther away from the ego vehicle. 
% Although some recently proposed methods incorporate image features for long-range detections, they do not scale well computationally with range, or are limited by depth estimation accuracy.
Another related problem is the label imbalance for faraway objects, which inhibits the performance of Deep Neural Networks at long-range.
To address the above limitations, we investigate two ways to improve long-range performance of current LiDAR-based 3D detectors. First, we combine two 3D detection networks, referred to as range experts, one specializing at near to mid-range objects, and one at long-range 3D detection. To train a detector at long-range under a scarce label regime, we further weigh the loss according to the labelled point's distance from ego vehicle.
Second, we augment LiDAR scans with virtual points generated using \gls{mvp}, a readily available image-based depth completion algorithm. 
Our experiments on the long-range \gls{av2} dataset indicate that \gls{mvp} is more effective in improving long range performance, while maintaining a straightforward implementation. On the other hand, the range experts offer a computationally efficient and simpler alternative, avoiding dependency on image-based segmentation networks and perfect camera-LiDAR calibration.
\end{abstract}

\glsresetall

%%%%%%%%%%%%%%%%%%%%%%%%%%%%%
\section{Introduction} 
\label{sec:intro}
As autonomous vehicles continue to make significant strides toward widespread adoption, ensuring their safety and reliability remains a paramount concern. One crucial aspect is the range of perception, a vital capability that enables self-driving cars to perceive and understand their surroundings accurately. While the need for long-range perception is apparent in various driving scenarios, it becomes especially critical for heavy-duty trucks on highways, where vehicles encounter potential hazards at high speeds requiring fast reaction time. 

A standardized safety assurance framework has been provided by authors in~\cite{shalev2017formal} 
% which includes a mathematical formulation 
for determining a safe longitudinal distance, considering factors such as comfortable deceleration, vehicle speed, and reaction time. Based on their findings, 
%a perception range exceeding 300 meters is necessary 
%ideally it is important to have an object detection range exceeding 300 meters
an object detection range exceeding 300 meters is needed
to ensure a safe braking distance for autonomous trucks in highway scenarios. Yet, the problem of 3D object detection at such ranges remains largely unexplored. 
At present, no publicly available datasets cover such a large range or provide data recorded on trucks. In our study, we use the \gls{av2} dataset~\cite{wilson2023argoverse}, which provides annotations up to 225~\si{m}. We define objects lying beyond 100~\si{m} to be \textit{long-range}, and those within 100~\si{m} to be \textit{mid-range}. 

%although we would like to test on truck data but none is available ....
%Ideally, one would use data recorded onboard trucks, which allow for longer viewing distances due to their larger size compared to automobiles. However, at present no such datasets are publicly available. Alternatively, large datasets recorded by cars need to be used, but they contain limited statistics of labeled objects at long distances. In this article, we focus exclusively on long range (or alternatively far away) 3D object detection, where we define long range to be greater than 100 meters. In our study, we use the \gls{av2} dataset~\cite{wilson2023argoverse}, which provides annotations up to 225 meters.

%At present, there are no publicly available datasets that encompass data recorded on trucks or annotated at such ranges. Alternatively, large datasets recorded by cars need to be used, but they contain limited statistics of labeled objects at long distances. In this article, we delve into typical

\begin{figure*}[!t]
     \centering
     \begin{subfigure}[b]{0.44\textwidth}
         \centering
        \includegraphics[width=\textwidth]{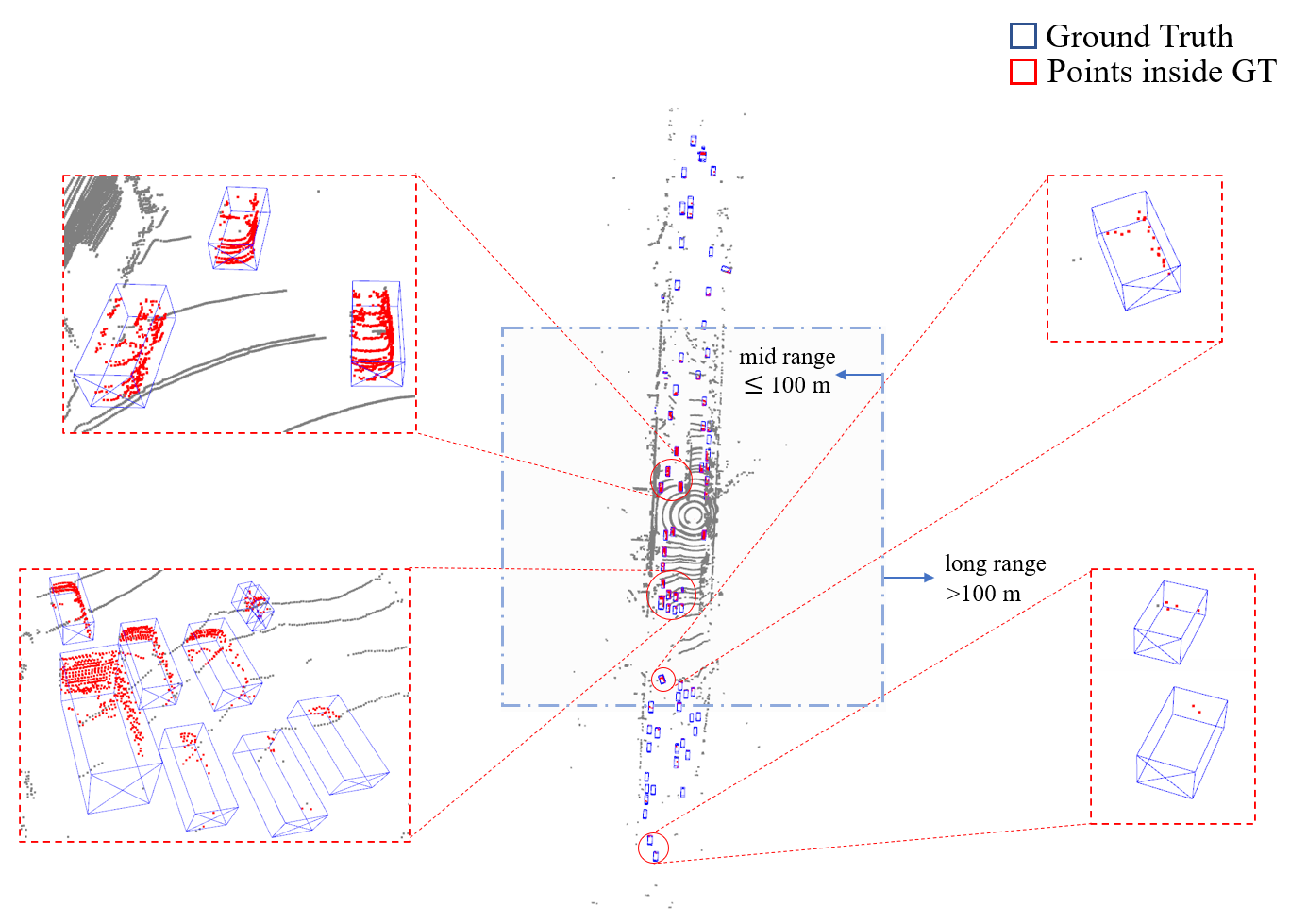}
        \caption{\gls{lidar} sparsity with range}
        \label{fig:overview_1}
     \end{subfigure}
     \hfill
     \begin{subfigure}[b]{0.49\textwidth}
        \centering
        \includegraphics[width=\textwidth]{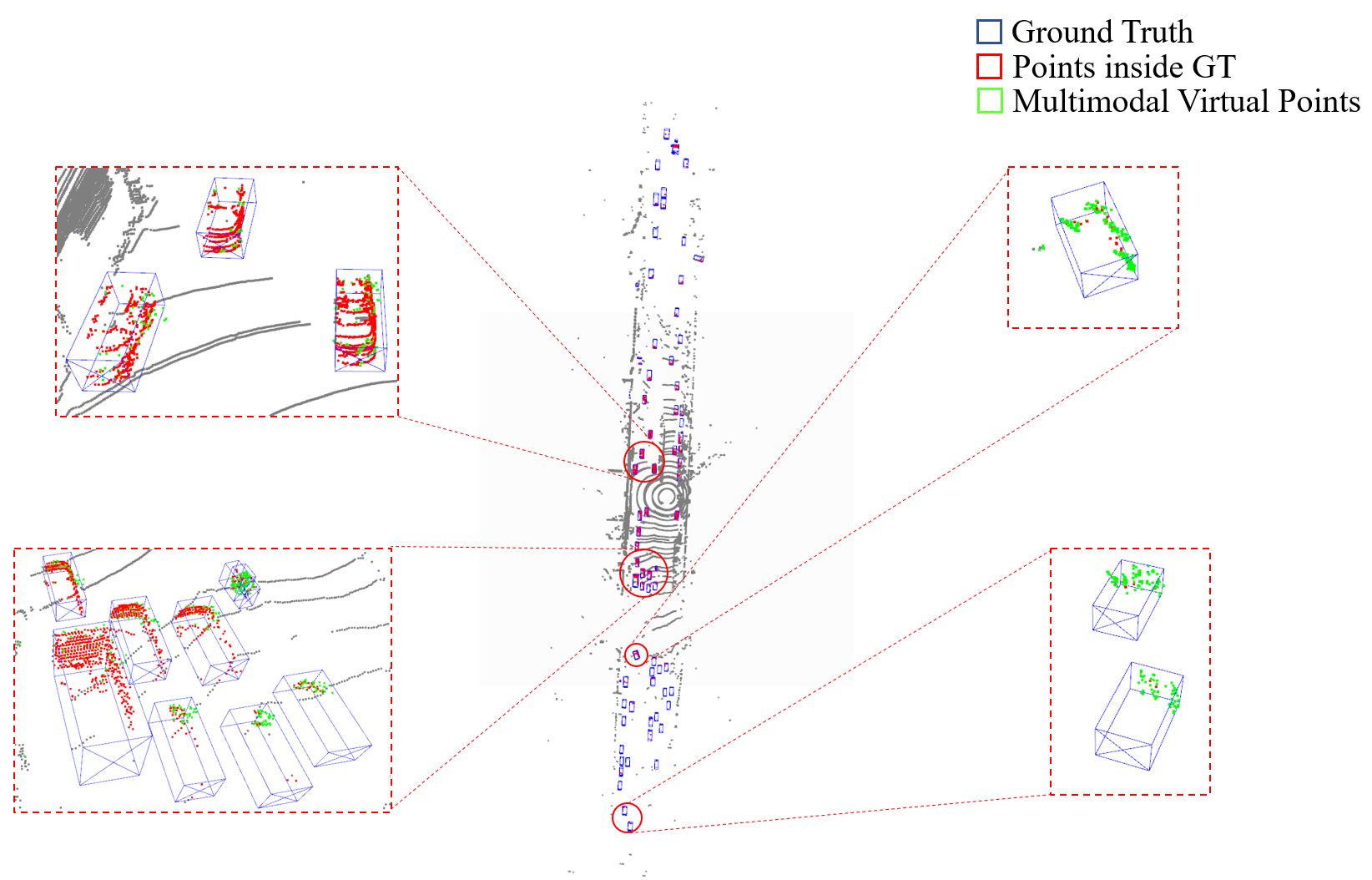}
        \caption{\gls{lidar} points (red) upsampled using virtual points (green)}
        \label{fig:overview_2}
     \end{subfigure}

     \begin{subfigure}[b]{0.49\textwidth}
         \centering
        \includegraphics[width=\textwidth]{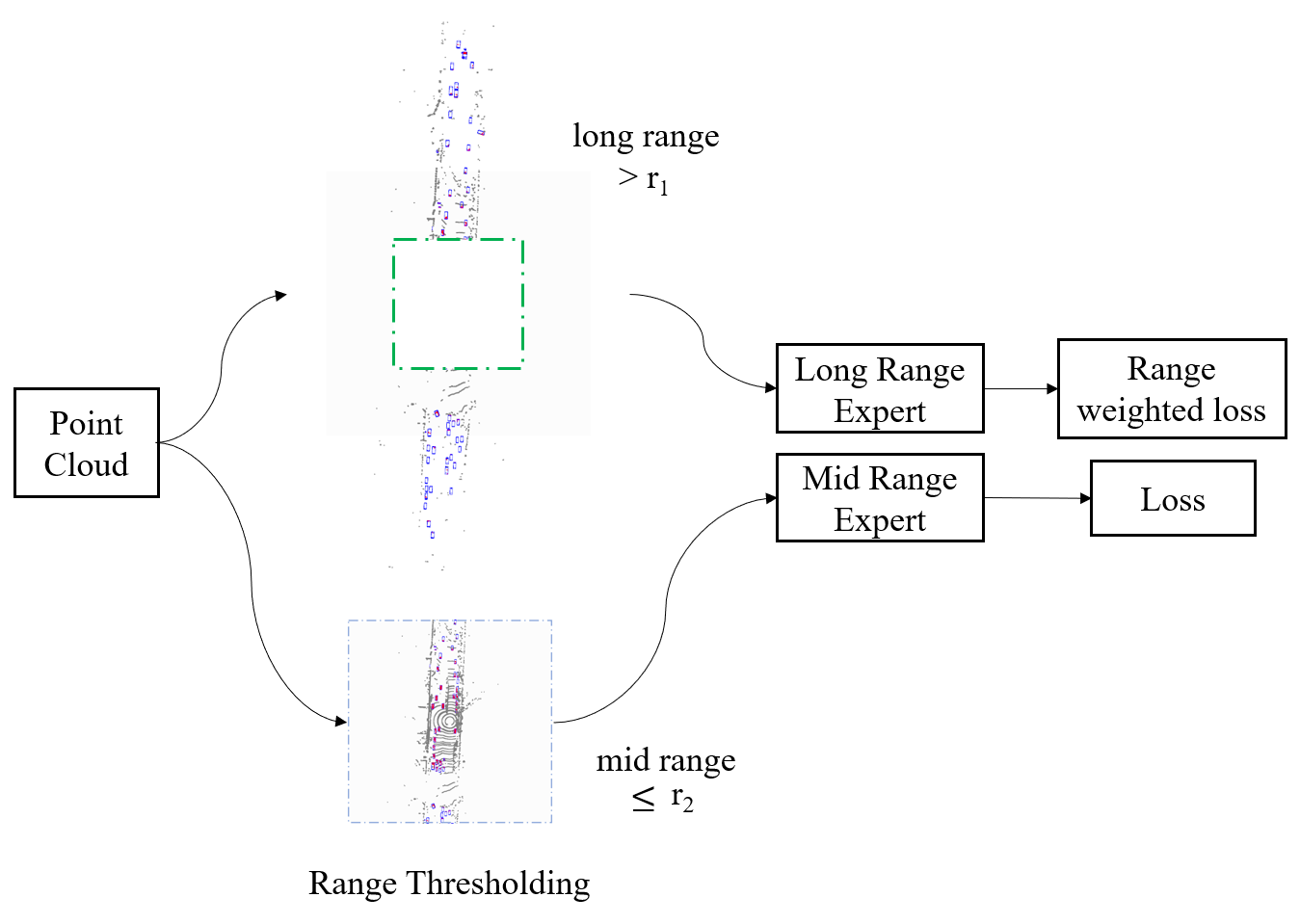}
        \caption{Range expert training}
        \label{fig:overview_3}
     \end{subfigure}
     \hfill
     \begin{subfigure}[b]{0.49\textwidth}
        \centering
        \includegraphics[width=\textwidth]{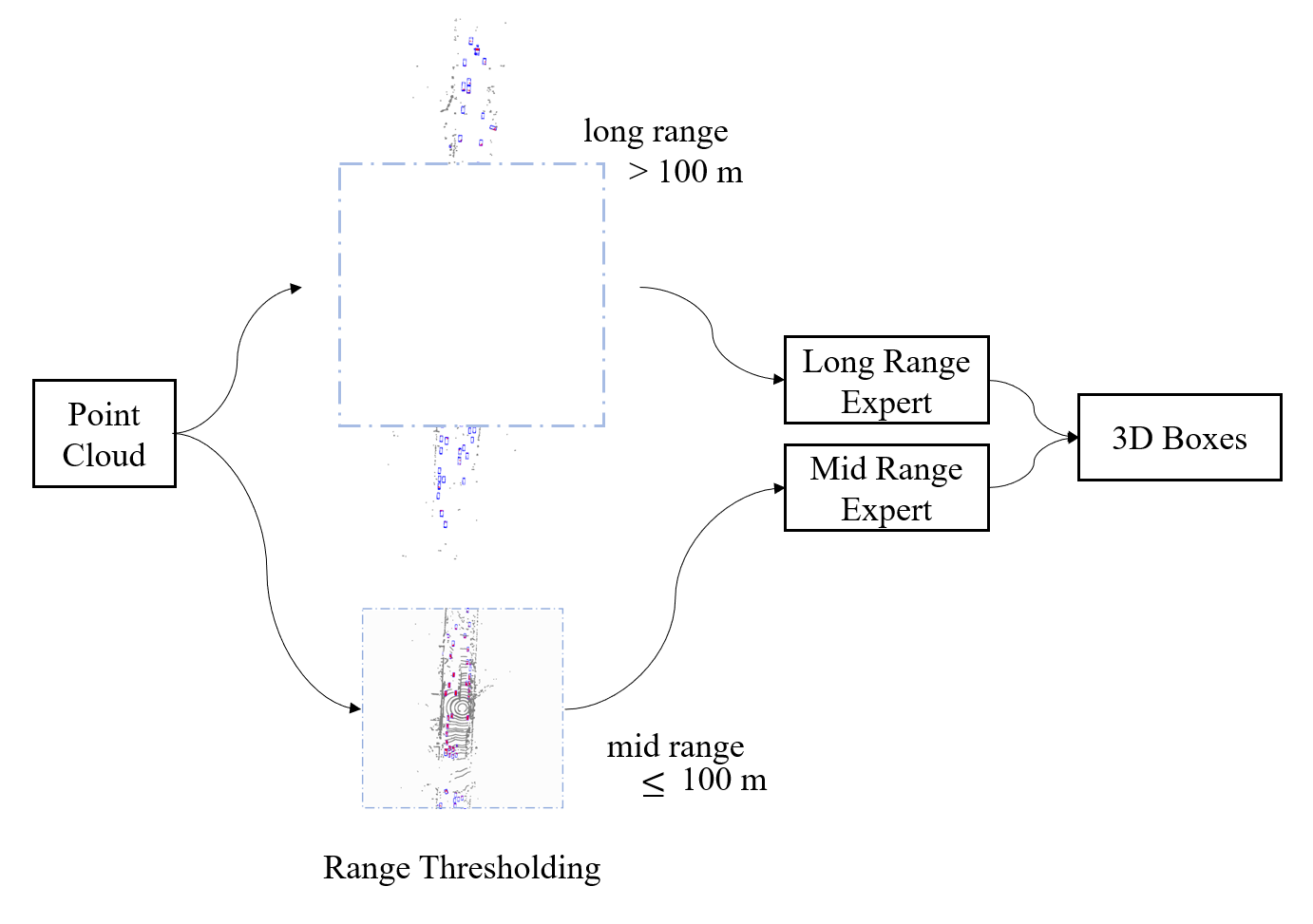}
        \caption{Range expert inference}
        \label{fig:overview_4}
     \end{subfigure}
     \caption{Increased \gls{lidar} sparsity with distance from ego vehicle is one of the major challenges facing long range (>100~\si{m}) 3D object detection. (\textbf{a}) shows that the two vehicles parked far away from the ego vehicle have fewer \gls{lidar} points, as well as appear different compared to those parked nearby. To address this problem, we analyze two approaches: (\textbf{b}) adding virtual points during training, and (\textbf{c}) training two neural networks, one specializing in mid-range object detection, and one for detecting long-range objects. 
     %To train a long range detector, we perform range masking, i.e. we remove points below a certain range $r$ (green box), including some but not all the mid range data.
    The parameters $r_1,r_2 \in [0,250]$, represented by the green and blue box respectively, delimit a subset of points used for training the long and mid-ranged experts respectively. Mid-range expert uses points within a distance $r_2$ of the ego vehicle, while points outside the distance $r_1$ are used to train the long-range expert. 
     %The green box delimits a subset of points which one may choose to train a specific range expert, i.e. a long (short) range expert is trained using points beyond (within) a distance $r$ of the ego vehicle.
     %The mid (long) range experts are trained using a subset of points with $r$
     %Secondly, we assign a larger weight to the loss component of the faraway object. 
     %Typically, choosing $r$=50 removes the domain gap to a certain extent, while providing enough points for the detector to generalize at long range. 
     (\textbf{d}) During inference, we combine outputs from the two networks, such that the long-range detector contributes detections beyond blue box (>100~\si{m}), and the mid-range detector contributes detections within blue box.  }
     \label{fig:overview}
\end{figure*}

At present, the realm of 3D object detection is largely led by methods relying on \gls{lidar} technology, as it excels in delivering precise point-based estimations of the environment. While state-of-the-art methods have showcased notable results in the context of mid-range detection, they often do not discuss findings for long-range detection or, when presented, exhibit poor performance in such scenarios~\cite{li2023fully,zhang2021faraway,gupta2023far3det}. We identify two primary issues that are currently constraining existing methods. 
%%%%%%%%%%%%%%%%%%%%%%%%%%%%%%%%%%%%%%%%%%%%%%%%%%%%%%%%%%%
The first issue is that the \gls{lidar} point cloud gets increasingly sparse with distance, as shown in~\Cref{fig:overview_1}. 
%An object far away from the ego vehicle to appears very differently compared to the one nearby. 
This causes a form of domain gap, which has been studied in the context of sim2real or sensor-to-sensor mapping applications~\cite{triess2021survey}, but not with regards to long range detection. Studies have suggested relying on camera-based methods for object detection beyond a specific range~\cite{gupta2023far3det}. Alternatively, 
%they have proposed the 
utilization of a dedicated network designed for processing long-range point data has been proposed~\cite{zhang2021faraway}.
Another consequence of \gls{lidar} sparsity is that a \gls{lidar} based detector lacks the necessary information to make precise predictions.

\begin{figure}[b!] 
\setlength\fwidth{0.8\linewidth}
\centering
\definecolor{mycolor1}{rgb}{1.00000,0.25000,0.15000}%
\begin{tikzpicture}

\begin{axis}[%
width=0.951\fwidth,
height=0.65\fwidth,
at={(0\fwidth,1.1\fwidth)},
scale only axis,
xmin=-12.5,
xmax=262.5,
xtick={  0,  50, 100, 150, 200, 250},
xlabel style={font=\color{white!15!black}},
xlabel={Distance~\si{(m)}},
ymin=0,
ymax=3800000,
ylabel style={font=\color{white!15!black}},
ylabel={\# Labeled Objects},
axis background/.style={fill=white},
xmajorgrids,
ymajorgrids,
legend style={at={(0.30,0.97)}, anchor=north west, legend cell align=left, align=left, draw=white!15!black}
]
\addplot[ybar interval, fill=white!60!black!, fill opacity=0.7, area legend] table[row sep=crcr] {%
x	y\\
0	3632022\\
50	2014961\\
100	714804\\
150	192465\\
200	21532\\
250	21532\\
};
\addlegendentry{All Classes}

\node[above, align=center, inner sep=0]
at (axis cs:25,250000) {55.2\%};
\node[above, align=center, inner sep=0]
at (axis cs:75,250000) {30.6\%};
\node[above, align=center, inner sep=0]
at (axis cs:125,250000) {10.9\%};
\node[above, align=center, inner sep=0]
at (axis cs:175,250000) {2.9\%};
\node[above, align=center, inner sep=0]
at (axis cs:225,250000) {0.3\%};
\addplot [color=black, dashed, line width=2.0pt]
  table[row sep=crcr]{%
100	0\\
100	2500000\\
};
\addlegendentry{Long Range ($>$ 100~\si{m})}

\end{axis}
\end{tikzpicture}%
\caption{Number of labeled objects for all categories as a function of range for the training set of the \gls{av2} dataset. The first two bins, which we refer to as mid-range, contain 85.8\% of the events in the data set. The 100-150~\si{m} range contains most of the long range objects, 10.9\%, leaving only 3.2\% objects over 150~\si{m}.}%\vspace{-1cm}
\label{fig:Statistics_dataset}
\end{figure}
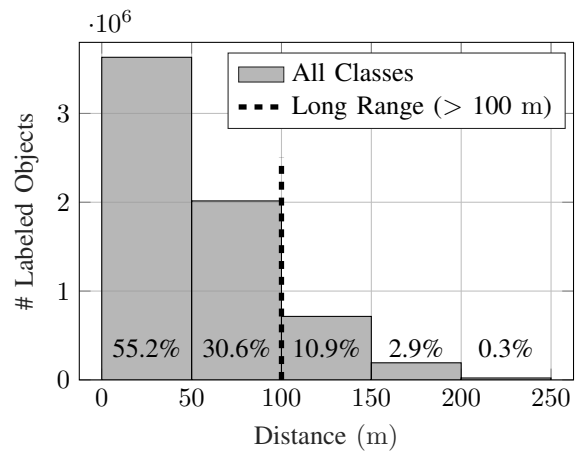 

The second issue is related to the uneven distribution of labeled objects across different ranges for training. Partially as a result of increasing \gls{lidar} sparsity with distance, there are fewer annotations at long-range, as shown in~\Cref{fig:Statistics_dataset}. Although many new datasets have been released covering long-range annotations partially,
%currently available open datasets 
they do not specifically focus on improving label imbalance for long-range annotations~\cite{caesar2020nuscenes, sun2020scalability, matuszka2022aimotive, alibeigi2023zenseact}.

% These shortcomings underscore the significance of our contributions:
In this work, we present the first comprehensive evaluation of 3D object detectors for distances beyond 100 meters, addressing a critical gap in the literature. Our novel contributions include:

\begin{itemize}
    % \item To our best knowledge, we are the first to provide a comprehensive evaluation of 3D object detectors beyond 100 meters range.
    % \item We investigate two ways to enhance long-range 3D object detection performance: Firstly we leverage recent developments in image based depth completion, and generate virtual points for objects selected by an image based segmentation network, as shown in~\cref{fig:overview_2}. An increased density of points is found to benefit both mid and long-range detection results.
    \item The integration of recent advancements in image-based depth completion to enhance long-range 3D object detection. We generate virtual points for objects identified by an image-based segmentation network, as illustrated in~\cref{fig:overview_2}. The increased density of points significantly improves the detection performance at both mid and long ranges.
    % \item Secondly, we train two networks as range experts, one specializing on mid range, and one learning long-range detections. During inference, we combine outputs from these networks such that they only detect objects in their respective range domains. (an overview of our proposed detection pipeline is shown in \cref{fig:overview_3,fig:overview_4}).
    \item The development and training of two specialized networks designed as range experts: one focused on mid-range and the other on long-range detections. During inference, we merge their outputs, ensuring each network operates within its optimal detection range. This approach is detailed in our proposed detection pipeline (refer to \cref{fig:overview_3,fig:overview_4}).
    % \item To tackle the challenge of label imbalance, we propose to weigh the labels according to their range.
    \item A strategy to address label imbalance by weighting labels based on their distance from the ego vehicle. 
    During training, higher weightage is assigned to the loss component of a faraway labeled object compared to the one nearer to the ego vehicle. This weighing is inversely proportional to the number of annotations in a particular range bin. 
    % \item We compare the performance of range expert with a detector trained with virtual points on the long-range \gls{av2} dataset.
    \item Moreover, we conducted a comparative analysis of our range expert networks against a detector trained with virtual points using the long-range \gls{av2} dataset.
    We also compare our best performing network to other state-of-the-art methods, observing significant improvements.
     % To our best knowledge, we are the first to provide a comprehensive evaluation of 3D object detectors beyond 100 meters range.
     % \item We evaluate the impact of adding MVP generated virtual points during training, as well as the range expert design, on the long-range \gls{av2} dataset.
    % \item Through extensive experiments, we show that range experts can be further improved when combined with image based depth completion methods that generate virtual points for objects selected by an image based segmentation network.
     % Although depth completion reduces the domain gap between mid-range and long-range objects, it might not remove it completely. To further address this issue, 
    % To mitigate the problem of \gls{lidar} sparsity,
    % \item To mitigate the problem of \gls{lidar} sparsity, 
\end{itemize}

The rest of this article is structured as follows. \Cref{sec:rel_work} describes the related work in the field of 3D Object Detection, with a focus on long range detection. \Cref{sec:method} describes range experts and outlines their training scheme for long range detection. \Cref{sec:expt} presents the experimental results and discusses their implications, while limitations, concluding remarks and future work are discussed in \Cref{sec:limitations} and \Cref{sec:conclusions}.

\section{Related Work}  \label{sec:rel_work}
This section presents the state-of-the-art in long-range 3D object detection and image based depth completion, as well as briefly describes the methods in fully sparse detection, most related to our work.

% \subsection{3D Object Detection}
% A variety of methods try to address the challenge of \gls{lidar} sparsity for 3D object detections.
% %this challenge 
% % The \gls{bev}-based methods try to address this challenge by employing
% The \gls{bev}-based methods employ multiple convolutional layers on the \gls{bev} feature map~\cite{lang2019pointpillars, zhou2018voxelnet}. As a result, the dense feature grid grows quadratically in memory with range~\cite{fan2022fully}. The point-based methods leverage sparsity to extend their range~\cite{zhang2022not} \cite{fan2022fully}, yet this advantage also results in a performance decline at longer distances. In recent years, progress has been made toward developing multimodal fusion based methods, to complement the sparsity. 
% %The most common approach relies on transforming dense multi-view camera features to \gls{bev} space~\cite{philion2020lift} and fusing with \gls{lidar} features using concatenation, addition, seqeeze and excitation networks~\cite{wozniak2023towards} or attention~\cite{bai2022transfusion}. 
% However, this relies on transforming 2D image features to 3D, which adds further computational cost~\cite{liu2023bevfusion, li2023dfa3d}. Moreover, noisy depth estimation from the camera limits the viability of feature fusion at long range~\cite{jiao2023msmdfusion}. 

\subsection{Long Range 3D Object Detection}
Point based or grid based methods utilizing sparse convolutions~\cite{yan2018second, graham2017submanifold} present interesting opportunities for long range 3D detection. VoxelNext~\cite{chen2023voxelnext} adds downsampling layers to existing 3D sparse convolution blocks to enlarge the receptive field, followed by pruning voxels with low feature magnitude, striking a balance between performance and computation.
IA-SSD~\cite{zhang2022not} first estimates per-point semantics and a centroid aware feature indicating closeness to a box center, and uses it to sample top $k$ points. 
%Thereafter a VoteNet~\cite{ding2019votenet} like module is used for feature extraction and box prediction per point.
While both methods demonstrate generally good performance, the feature pruning might remove information for long range objects.

Recently, there has been growing interest in multimodal fusion based methods for long range detection. In Far3Det~\cite{gupta2023far3det}, the authors report that fusion based methods outperform  \gls{lidar} only methods, given that the inaccuracies in camera based depth estimation are taken into account during evaluation. They propose range adaptive NMS thresholds and a straight-forward late fusion scheme, where camera-based detections are preferred beyond certain range. However, at present many camera based methods for 3D detection rely on converting image features into \gls{bev} representations, which poses significant computational challenges for long range~\cite{jiang2023far3d}. %Noting that learning based methods don't generalize well for sparse faraway points, 
Zhang et. al.~\cite{zhang2021faraway} classify and cluster the points inside frustum obtained from image based segmentation networks, and design a novel FFNet to regress the 3D bounding box. \gls{fsf}~\cite{li2023fully} similarly uses image to cluster \gls{lidar} points. They further utilize the \gls{fsd}~\cite{fan2022fully} network to filter noisy points at mask boundaries. The image segmented cluster is fused with \gls{lidar} segmented clusters for feature extraction and box prediction. To the best of our understanding, none of the methods listed above discuss or present results specifically at long range, i.e. above 100~\si{m} distance.

% \textcolor{blue}{I need one sentence after each section in SoA, what they have missed, it is not clear e.g. in A what they have not done, and what you will do, you can just say, none of the methods show or discussed results for above 100 m , you need to clarify they miss something that's why you have this article}

\subsection{Fully Sparse Detectors}
%Point based 3D detectors typically rely on computationally expensive search queries for defining neighborhoods and often assign a point to multiple clusters for multiscale refinement~\cite{qi2017pointnet++}. As a result, methods in this area struggle in terms of scalability and often have to subsample the point cloud~\cite{mao20223d}.
\gls{fsd}~\cite{fan2022fully} is among the first computationally efficient point based methods. It overcomes the drawbacks of neighborhood querying and multiple cluster assignments faced by previous methods, by first pre-segmenting and voting on the point cloud using a SparseUNet~\cite{graham2017submanifold}. 
%Each segmented point also votes for an object center.
Clustering~\cite{he2017connected} on these voted centers assigns a unique grouping to every point, avoiding ambiguity. This is followed by designing a novel \gls{sir} module to extract cluster features and predict boxes.
%\gls{fsd} \cite{fan2022fully} also pre-segments the point cloud, but furthermore does clustering \cite{he2017connected} on voted centers, followed by designing a novel \gls{sir} module to extract cluster features and predict boxes. 
%The authors also propose an optional refinement stage where residuals for box proposals from previous stage are estimated. 
%\cite{fan2023super} incorporates time-sequential LiDAR information in \gls{fsd}, by only including points with significantly new information (called residual points). To account for noise in clustering, 
%FSDv2 \cite{fan2023fsdv2} instead voxelizes the voted centers to form virtual voxels, and fuses it with features from segmentor and real voxel for refinement. 

To account for noise in clustering, \gls{fsd}v2~\cite{fan2023fsdv2} instead voxelizes the voted centers to form \textit{virtual voxels} to capture the ambiguity/noise in clustering. Thereafter multi-scale features from a sparseUNet decoder are fused with those from virtual voxels and real voxels (obtained from the foreground segmented points). 
%This feature interaction proves beneficial, as demonstrated by further improvement in detection performance on AV2 dataset.
We use \gls{fsd} and \gls{fsd}v2 as backbones for mid range and long range experts, due to their state-of-the-art results on long range 3D object detection, as well as their computational efficiency.
% Our method further improves the performance of these models, making them, to our best knowledge, the current state-of-the-art results with improvements both at mid and long ranges.
% \textcolor{blue}{"results on long range 3D object detection" you need to be clear what you have added to FSD, what you wrote is FSD already solve it efficienty. \textbf{FIXED. PLS RECHECK}}

\subsection{Image Based Depth Completion}
Rich semantic information from image based 2D detectors can provide valuable cues for long range objects, such as rough estimation of object center~\cite{zhang2021faraway, qi2020imvotenet}, enhanced feature representations~\cite{vora2020pointpainting, xu2021fusionpainting}. An intriguing research direction, which directly targets the \gls{lidar} sparsity problem is image-based depth completion. Prior research has explored this approach to a lesser extent in the context of long-range 3D detection. In the work by Zhang et al.~\cite{zhang2018deep}, a \gls{dnn} is used to estimate surface normals and occlusion boundaries, and combine them with raw depth measurements for depth completion. The work presented in~\cite{ku2018defense} uses classical image processing operations like dilation with custom designed kernel, hole filling, and median filtering to fill the depth map. 

The method \gls{mvp}~\cite{yin2021multimodal} first uses an image based instance segmentation network to generate instance masks where $s$ pixels are randomly sampled without repetition for each mask. Next, for each pixel, the nearest \gls{lidar} point, obtained by projecting the point cloud onto mask using intrinsic and extrinsic parameters, is located. The nearest \gls{lidar} points are then used to assign depth to these $s$ pixels, which are lastly unprojected back to 3D to obtain the virtual points. Among existing approaches, the \gls{mvp} stands out as one of the top-performing methods, notable for its efficacy and simplicity. 
\section{Methodology} \label{sec:method}
% In order to improve performance on long-range points, 
% %we use a combination of a mid range and a long range expert, and in the proceeding section, we investigate ways of training it. 
% two different approaches are used: range thresholding and range weighing. To address the \gls{lidar} sparsity problem, we also evaluate the impact of \gls{mvp} to generate virtual \gls{lidar} points at all ranges and combine them with real points for training. Finally, we combine our best performing range experts into our final state-of-the-art model.
%%The addition of virtual points is beneficial for both mid range and long range performance,  
Due to the increase in sparsity of the LiDAR point cloud as a function of the distance, objects at different distances appear to be radically different. To tackle this issue, we investigate two approaches: Firsty, we use \gls{mvp}, as described in previous section, to generate virtual points. These virtual points are then fused with the \gls{lidar} data before training FSD and FSDv2.
Second, we train two specialized versions of the detector, each optimized for a specific distance domain: one for mid-range and another for long-range points. This specialization is achieved through range thresholding, which allows each detector to focus on learning features relevant to its designated range domain. 
The long-range detector, in particular, employs a weighted loss function that accounts for the object's distance.
During inference, we employ a late fusion strategy to integrate the outputs from both range experts. 
The following sections delve into the detailed implementation of our range experts, their training process, and the fusion mechanism

% We take a baseline 3D object detector, and train two copies of it: one for mid ranged points, and the other for long ranged points.
% We use range thresholding to train 3D object detectors within a certain range domain. Particularly for long range detector, we weigh the loss function according to the object's range. Lastly, during inference we combine the output from two range experts using late fusion.

%\textcolor{blue}{A weighted loss function that accounts for the object's distance, referred to as range weighing was also tested as an alternative to range thresholding. The range experts performed better.}

\subsection{Notation} \label{subsec:notation}
The set $\mathcal{P}=\{(x_i, y_i, z_i, f_i),i\in\mathbb{N}\}$ represents a point cloud in $\mathbb{R}^4$. 
%In this article, we omit the time dimension as we consider a single point cloud. 
%We define the range $r$ of an object as the maximum distance between the ego vehicle and the object center in the x or y plane, such that the set of points with $r = x$~\si{m} draw a square around the ego vehicle with length of $2x$.   
% We define $g(p) = \max(\lvert x\rvert, \lvert y \rvert)$ as a function that computes the range of a point $p\in\mathcal{P}$, and $\lvert.\rvert$ present absolute value. 
% A range expert is denoted by F$_{\theta, r_1-r_2}$ where F$_{\theta}$ (or F for brevity) is the \gls{dnn}, parameterized by $\theta$ and chosen as the core model, 
Range experts are denoted by F$_{r_1-r_2}$ where F is the \gls{dnn} used for training, 
and $r_1, r_2 \in \mathcal{R}=\{0, 50, 100, 150, 200, 250\}$ are the range thresholds (in meters) within which it is trained. 
For a range threshold $r$, we consider a square axis aligned region where $r$ denotes the distance from the origin to the border along the x or y axis.
The set $\mathcal{R}$ further defines a set of bins $\mathcal{B}$ separated by fixed length interval of 50~\si{m}.
% , with the range thresholds $r_1$ and $r_2$ acting as the edges for binning. For example, the network $F_{100-250}$ would cover three range bins (including the lower bound but excluding the upper bound): 100-150~\si{m}, 150-200~\si{m}, and 200-250~\si{m}.
The maximum range is set to 250~\si{m} for the \gls{av2} dataset. Specifically, F$_{r_1-r_2}$ is trained to perform inference for points $p\in\mathcal{P}$ such that $r_1 \leq r \leq r_2$.
% , and we investigate which ranges should be used for training. 
In this paper, we consider F$_{0-250}$ as the baseline model upon which we build our contributions. 
Furthermore, we denote F$_{r_1-r_2}^w$ as an expert trained using range weighing.
%
% % Given a set of points $\mathcal{P}$, we define a function $g(p)$ that computes the range of a point $p=(x, y, z, f)$, like so:
% % \begin{gather}
% %     g(p) = max(abs(x, y))
% % \end{gather}
% Given a set of points $p \in \mathcal{P}$, we first denote $g(p)= max(abs(x, y))$ as a function that computes the range of a point $p=(x, y, z, f)$.
% Our design consists of two range experts, denoted by $F_\theta\_e@r$. $F_{\theta}$ (or $F$ for brevity) is a \gls{dnn} chosen as the code model, $e \in \mathcal{E} = \{mid, long\}$ denotes the range expertise and $r \in \mathcal{R} =  \{0, 50, 100, 150, 200, 250\}$ denotes the thresholding for range masking. 
% Concretely for range threshold $r$, the mid range network is trained for all points $p$ such that $g(p) \leq r$, and the long range network is trained for points $p'$ such that $g(p') > r$.
% %for which $g(p)$ is less than or equal to r, and the long range network is trained for points with $g(p)$ greater than $r$.
% %Concretely, for $r  \in \mathcal{R}$, the mid range network is trained for points $\mathcal{P}' = p' \in \mathcal{P}, \text{ such that } p' \leq r'$, and the long range network is trained for points $\mathcal{P} - \mathcal{P}'$.
% Hence $F\_long@0$ is the same as $F\_mid@250$ for a dataset with maximum range less than 250 meters. 
% % The choice of two range experts is motivated by simplicity, and in practice one can choose more than two range experts, albeit at an additional computational cost, as discussed in \textbf{RUNTIME SUBSECTION}.
%
\subsection{Range Thresholding}  \label{subsec:range_experts}
%Given the large difference in the number of labeled objects as a function of their distance, which was illustrated in \cref{fig:Statistics_dataset} 
% Due to the increase in sparsity of the LiDAR point cloud as a function of the distance, objects at different distances appear to be radically different.
%far away objects look radically different to nearby objects. 
% We use range thresholding to target objects in different domains. 
%and therefore maximise the performance of our model across $\mathcal{R}$.
%Range experts exploit this domain gap of the objects as a function of the distance to improve performance of points that lay in a specific range. %Given the domain differences as a funcion of the distance, it is only natural that models such as FSD are especially good at detecting nearby objects within distances below 50 m, while their performance degrades with the distance.
%This is achieved by 
Range thresholding consists of training the network using only a subset of 
% objects 
points within a specific range interval $r_1-r_2$.
% we are aiming to target. 
The resulting networks are referred to as range experts.
% In this paper, we compare range experts specialised on object detection at mid and long range with those trained on the full dataset, hereby referred to as general.
%Furthermore, we combine mid and long range experts to maximise the performance of our models. 
Given the large difference in the number of labeled objects as a function of their distance, which was illustrated in \cref{fig:Statistics_dataset}, a mid-range expert is not crucial since the baseline, F$_{0-250}$, already performs fairly well as a mid-range detector, however, specialised long-range experts are clearly required.  %but using the baseline as a long range expert wastes a lot of computation, as all the detections between 0-100~\si{m} would need to be discarded. 
We evaluate the performance of the long-range experts as a function of the range thresholding values. 
%, namely FSD$_{50-250}$, FSD$_{100-250}$, i.e. discarding points with range below 100 meters, 
Similarly to \cite{peri2023empirical}, which tests range thresholds on \gls{bev} methods,
we observe that as we increase the lower bound for range thresholding, the network performs poorer compared to baseline at long range. This is because, as seen in \cref{fig:Statistics_dataset}, we end up discarding larger and larger proportion of labels while training, and the network does not have enough data to learn from. This problem inspired us to also implement the range weighing. 

\subsection{Range Weighing}    \label{subsec:range_weighing}
Range weighing is aimed at compensating for the imbalance of labeled objects at large distances, which was illustrated in \cref{fig:Statistics_dataset}. This is achieved by increasing the impact of objects that are further away on the loss function. In order to strike a balance between range thresholding and retaining sufficient labels for training, we propose range adaptive weighing. The loss corresponding to a predicted cluster/virtual voxel is weighed according to its range. In general, the label weight $w_b$ for the range bin $b$ is given by:
\begin{gather}
    w_b = \frac{N}{n_b \times B}
\label{eq:weights}
\end{gather} 

\noindent where $N$ is the total number of labels for all classes in the dataset, $n_b$ is the number of objects in range bin $b$ and $B$ is the number of bins covered by the range expert. 
% ~\Cref{table:range_wts} denotes the range weights for both mid and long range experts for different values of range thresholding $r_1$ and $r_2$. These weights are computed according to bin-wise label statistics (as shown in \cref{fig:Statistics_dataset}) for the training set. 
The weight $w_b$ is applied to the Focal Loss \cite{lin2017focal} and $L_1$ Loss \cite{girshick2015fast}, which are used as classification and regression losses for the prediction head for bounding boxes. 
%Specifically, given $Q$ predicted clusters (or virtual voxels) and $L$ matched ground truth targets, the losses are modified as: 
%
%\begin{gather}
%    L_{cls}^* = FocalLoss(C, L, w_b, \gamma, \alpha,)\\
%    L_{reg}^* = L_1Loss(C, L, w_b)
%\end{gather}
%where $\gamma$, $\alpha$ are hyperparameters. 
%, and denoting $w_b=1$ gives the original loss
The other loss terms remain unchanged (refer to \cite{fan2022fully}, \cite{fan2023fsdv2} for more details). 
% We denote F$_{r_1-r_2}^w$ as an expert trained using range weighing.
The range weights for both mid and long range experts for different values of range thresholding $r_1$ and $r_2$ are denoted in ~\Cref{table:range_wts}.

\begin{table}[h] 
\centering 
\scriptsize
% \footnotesize 
\caption{Range weights for different values of range thresholds. $r_1$ and $r_2$ are lower and upper limits for thresholding.} 

\renewcommand{\arraystretch}{1.2}
\setlength{\tabcolsep}{3pt}\begin{tabular}{cc|ccccc}
 \hline 
  $r_1$  & $r_2$ & 0-50 & 50-100 & 100-150 & 150-200 & 200-250 \\
 \hline
 0 & 250 & 0.360 & 0.661 & 1.852 & 6.486 & 52.332\\
 50 & 250 & 0 & 0.367 & 1.030 & 3.607 & 29.103\\
 100 & 250 & 0 & 0 & 0.440 & 1.542 & 12.440\\
 %150 & 250 & 0 & 0 & 0 & 0 & 0\\
 %0 & 100 & 0.772 & 1.418 & 0 & 0 & 0\\
 \hline 
\end{tabular} 
\label{table:range_wts}
\end{table}

% We combine them into a single network, which we refer to as \textit{RangeF}.

\subsection{Late Fusion}    \label{subsec:late_fusion}
We studied various combinations of range thresholding and range weighing
% %Section~\ref{subsec:ablation_studies}, 
% in \Cref{subsec:range_thresh_and_wt},
using FSD~\cite{fan2022fully}, and chose F$_{0-100}$ and F$^w_{50-250}$ as the mid and long-range experts.
%, where F$^w$ denotes range wighted model. 
% \textcolor{red}{We finetune our models using virtual points generated by \gls{mvp} for these configurations and combine them into a single network, as shown in \cref{fig:overview_2}. Consequently, FSD$_{0-100}$ and FSD$_{50-250}$ form the mid and long-range experts of a single network, which we refer to as \textit{RangeFSD}. Similarly, we obtain another network \textit{RangeFSDv2}, with FSDv2~\cite{fan2023fsdv2} as backbone. }
During inference, the detections from the mid and long-range experts are combined such that
%they contribute boxes in their specific ranges only . Specifically, 
the mid-range expert contributes detections with centers from 0-100 meters, whereas the long-range expert contributes detections from 100-250 meters. 

\section{Experiments}   \label{sec:expt}

A variety of experiments and ablation studies are documented in this section. We assess the impact of our individual contributions and compare the best performing model to other state-of-the-art methods from the literature.

\subsection{Experimental Setup}  \label{subsec:expt_setup}
% \textbf{Dataset:} Argoverse 2 contains 1000 scenes
\textbf{Dataset:} We conduct experiments on the \gls{av2} perception dataset, which contains 1000 scenes divided into 700 for training and 150 each for validation and testing. Each scene captures roughly 15~\si{s} of data on two 32-beam \glspl{lidar} at 10~\si{Hz} but with a 180$^\circ$ difference in spinning angle, seven color cameras and two forward-facing stereo cameras at 20~\si{Hz}, providing a surround view of the vehicle. Objects of interest are annotated at 10~\si{Hz} among 30 categories, up to 225~\si{m} away from the ego vehicle.

%\textcolor{red}{worth in the result you compare the quality of your query with the FSF method as we claim yours is simpler so it should be good enough too}

\textbf{Implementation Details:}
We base our implementation on \gls{fsd}~\cite{fan2022fully}, which is in turn developed using mmdetection3d library~\cite{mmdet3d2020}. We start with a pre-trained \gls{fsd} and train for six epochs using \gls{cbgs}~\cite{zhu2019class}.
Thereafter we train the mid and long-range experts for further 3 epochs and combine them, as explained in~\cref{sec:method}.
The combined network is referred to as range expert (or RE for brevity).
For \gls{mvp}, We use \gls{htc}~\cite{chen2019hybrid} pretrained on NuImages dataset as the default 2D image segmentor for generating the virtual points. 
% Due to less data at long range and avoid over fitting we reduce epics to three and continue the training
% Thereafter, we enable \gls{mvp} and train further using \gls{cbgs} for three epochs. 
Thereafter, we further train an FSD using these virtual points for 3 epochs with \gls{cbgs}. 
We use the same data augmentations and learning parameters as~\cite{fan2022fully} and train on eight Nvidia A-100 GPUs.
%For data augmentations while training, we use global random rotations between $[-\pi/4,\pi/4]$, global random scaling between . 
%We train on 8 Nvidia A-100 GPUs, 
%with batch sizes 16, 75 and 192 for FSD depending on the range thresholding configuration, and use batch size of 1 for evaluation.  

\textbf{Metrics:} \gls{av2} uses \gls{map} and \gls{cds} scores for evaluating 3D detection. In this work, we report the \gls{map}, which is defined as the mean of \gls{ap} across $C$ classes, where \gls{ap}$_j$ for class $j$ is the discrete integral of precision $p_d(a)$ interpolated at 100 discrete recall values $a\in [0,1]$, averaged across four chosen distance thresholds $d$ (in meters).
\begin{align}
    mAP &= \frac{1}{C}\sum^{C}_{j=1} AP_j \\ 
    AP_j &= \frac{1}{4}\sum_{d=\{0.5, 1, 2, 4\}} \frac{1}{100}\sum_{a=\{0,0.05,\ldots,1 \}} p_d(a)
\end{align}
%
%The \gls{cds} weights \gls{map} with average true positive scores, defined as the complement of one over normalized true positive errors for translation, scale, and orientation. For more details, please refer to \cite{chang2019argoverse}. 
Unless otherwise specified, we use an evaluation range of 0-250~\si{m} in our experiments.

% At medium or short range, \gls{cds} is beneficial as it provides a fine-grained idea of detection results. But at long range, the \gls{cds} and \gls{map} are likely to be correlated. Instead at long range, being able to detect all the true positives is of paramount importance. In doing so, if we get a few more false positives, that is acceptable. This intuition can be measured using the mean average recall (mAR). Concretely, we define mean average recall as:
% following pascal voc we keep 100 max detections. In our experiments we discard \gls{cds}, and instead look at the combination of mAP and mAR.   
\subsection{Performance Breakdown}  \label{subsec:performance_breakdown_1}
% Our contribution can be divided into two components: implementation of \gls{mvp} and range experts (referred to as RE for brevity). 
In order to assess the impact of range experts and \gls{mvp} towards enhancing 3D object detection, a detailed performance breakdown, shown in \Cref{tab:performance_breakdown_1_mAP}, is performed.  

\begin{table}[h] 
\centering 
\scriptsize
% \footnotesize 
\caption{mAP comparison (all classes). $\dagger$: provided by authors~\cite{fan2022fully}. *: re-trained by us. RE: Range Experts. MVP: Multimodal Virtual Points.} 
\label{tab:performance_breakdown_1_mAP}
\renewcommand{\arraystretch}{1.2}
 \setlength{\tabcolsep}{3pt}\begin{tabular}{l|c|ccccc}
 \hline 
  Method & 0-250 & 0-50 & 50-100 & 100-150 & 150-200 & 200-250  \\
 \hline
 FSD$\dagger$ &  28.5 & 43.7 & 12.6 & 3.3 & 1.0 & 0.3 \\
 FSD* &  29.8 & 44.9 & 14.3 & 4.1 & 1.0 & 0.3 \\ 
 FSD*+RE & 30.2 & 45.5 & 14.8 & 4.3 & 1.2 & 0.3 \\ 
 FSD*+MVP & 31.5 & 46.4 & 16.6 & 5.4 & 1.8 & 0.5\\
 FSD*+RE+MVP & 31.9 & 46.9 & 16.9 & 5.8 & 2.0 & 0.5 \\ 
 \hline 
\end{tabular} 
\end{table}

To guarantee a fair comparison, we re-train the baseline networks for six epochs using \gls{cbgs}, which we denote by $*$. We also provide the performance reported by the authors of the baseline for completeness, denoted by $\dagger$. For \gls{fsd}, we reduce the minimum points for clustering from two to one, to enable clustering at long range. It's observed that using \gls{cbgs} seems to slightly benefit long range performance.
% Improvements are observed for both RE and \gls{mvp}, but the latter is more effective overall.
Compared to FSD$^*$, RE and \gls{mvp} contribute to 0.4\% and 1.7\% improvement overall.
But it becomes difficult to access the performance beyond 100~\si{m} range. Upon analysing the class distribution across different range bins, we found that roughly 50-70\% objects belong to a single class category: vehicle. Especially beyond 100~\si{m}, other categories have very few annotations. Due to this their class-wise \gls{ap} falls close to zero, making the \gls{map} very low. 
In order to remedy this, we present the performance breakdown with only the \gls{ap} for vehicle class in table~\Cref{tab:performance_breakdown_1_AP_vehicle}.

\begin{table}[h] 
\centering 
\scriptsize
% \footnotesize 
\caption{AP comparison (Vehicle class only). $\dagger$: provided by authors~\cite{fan2022fully}. *: re-trained by us. RE: Range Experts. MVP: Multimodal Virtual Points.} 
\label{tab:performance_breakdown_1_AP_vehicle}
\renewcommand{\arraystretch}{1.2}
 \setlength{\tabcolsep}{3pt}\begin{tabular}{l|c|ccccc}
 \hline 
  Method & 0-250 & 0-50 & 50-100 & 100-150 & 150-200 & 200-250  \\
 \hline
 FSD$\dagger$ &  70.5 & 88.9 & 59.7 & 26.8 & 12.1 & 2.6 \\
 FSD* &  71.5 & 89.0 & 61.7 & 29.2 & 13.9 & 3.6 \\ 
 FSD*+RE & 71.9 & 89.0 & 62.5 & 30.0 & 15.6 & 3.4 \\ 
 FSD*+MVP & 74.5 & 89.0 & 66.0 & 39.2 & 23.2 & 6.0\\
 FSD*+RE+MVP & 74.5 & 89.2 & 66.0 & 39.3 & 23.2 & 6.0 \\ 
 \hline 
\end{tabular} 
\end{table}

% At long-range (i.e. beyond 100~\si{m}), the range experts enhance vehicle \gls{ap} by up to 1.7\%, while the \gls{mvp} demonstrates an even more substantial improvement, reaching up to 10\%.
% On the other hand, it must be noted that \gls{mvp} uses virtual points upsampling during evaluation as well.
% This large improvement can be explained by the fact that contrary to other approaches, \gls{mvp} uses virtual points upsampling during evaluation as well.
For distances exceeding 100 meters, range experts contribute to an increase of up to 1.7\% in vehicle \gls{ap}, whereas \gls{mvp} yields a significantly greater enhancement, achieving an increase of up to 9.3\%. This marked improvement is primarily due to \gls{mvp}'s application of virtual points upsampling during the evaluation phase, a technique not employed by other approaches.
Furthermore, MVP relies on precise camera-\gls{lidar} calibration. Any misalignments in the extrinsic calibration process can result in inaccuracies in depth completion and the potential exclusion of objects when nearby \gls{lidar} points are absent. This limitation is particularly notable for distant objects and can significantly impact the overall performance of the network. 

The integration of \gls{mvp} with range experts yields a modest improvement in overall \gls{map}, however, it appears to have a limited effect on enhancing vehicle AP. This outcome may stem from \gls{mvp}'s capability to bridge the domain gap between mid-range and long-range points, reducing the contributions of range experts.
Our results underscore the substantial room for
improvement within the current state-of-the-art for long-range 3D detection.

% for short range, but improvements to long range are negligible.

% RE and MVP give improvement and their combination improves things further.
% but it gets super hard to access performance beyond 100 m because numbers are in single digits. 
% upon analysing the bin wise class distribution we found that 50-70\% objects are of vehicle class. esp beyond 100 m, other categories have very few annotations. due to this their AP numbers are close to zero, bringing the mAP very low. to look at the long range performance more closely we present above table but only with AP for vehicles class in table XXX. readers are advised to see mAP till 100 m and vehicle AP beyond it.
% Range experts improve 0.8\% and 1.7\% in 100-150 m bin and 105-200m bin, whereas MVP improves by around 10\% for same bins. 
% Range experts improve upto 1.7\% vehicle AP at long range and 0.4\% mAP overall, whereas MVP improves upto 10\% for long range and 1.7\% overall. 
% notably, the combination of MVP and RE is slightly beneficial for short range, but improvements to long range are negligible. 

\subsection{State-of-the-art Comparison}    \label{subsec:sota}
% Points to cover here: fsdv2 with mvp and range weights gives best performance. although many categories improve,  

\begin{table*}[h!] 
\centering 
% \tiny
\scriptsize
\caption{Comparison to state-of-the-art on \gls{av2} validation split. The evaluation range is 0-200~\si{m}. $\dagger$: provided by authors of \gls{av2}~\cite{wilson2023argoverse}. $\ddagger$: reimplemented by \gls{fsd}~\cite{fan2022fully}. *: re-trained by us. RE: Range Experts. MVP: Multimodal Virtual Points~\cite{yin2021multimodal}. C-Barrel: construction barrel, MPC-Sign: mobile pedestrian crossing sign, A-Bus: articulated bus, C-Cone: construction cone, V-Trailer: vehicular Trailer, W-Device: wheeled device, W-Rider: wheeled rider.} 
 
\setlength{\tabcolsep}{2.3pt}\begin{tabular}{l|ccccccccccccccccccccccccccc} 
 
 % \hline \\
 %  Method & \rot{mAP } & \rot{Articulated Bus } & \rot{Bicycle } & \rot{Bicyclist } & \rot{Bollard } & \rot{Box Truck } & \rot{Bus } & \rot{C-Barrel } & \rot{C-Cone } & \rot{Dog } & \rot{Large Vehicle } & \rot{MBT } & \rot{MPC-Sign } & \rot{Motorcyle } & \rot{Motorcyclist } & \rot{Pedestrian } & \rot{Vehicle } & \rot{School Bus } & \rot{Sign } & \rot{Stop Sign } & \rot{Stroller } & \rot{Truck } & \rot{Truck Cab } & \rot{Trailer } & \rot{Wheelchair } & \rot{W-Device } & \rot{W-Rider }\\
 \hline \hline\\ 
  Method  & \rot{mAP} & \rot{Vehicle } & \rot{Bus } & \rot{Pedestrian } & \rot{Stop Sign } & \rot{Box Truck } & \rot{Bollard } & \rot{C-Barrel } & \rot{Motorcyclist } & \rot{MPC-Sign } & \rot{Motorcycle } & \rot{Bicycle } & \rot{A-Bus } & \rot{School Bus } & \rot{Truck Cab } & \rot{C-Cone } & \rot{V-Trailer } & \rot{Sign } & \rot{Large Vehicle } & \rot{Stroller } & \rot{Bicyclist } & \rot{Truck } & \rot{MBT } & \rot{Dog } & \rot{Wheelchair } & \rot{W-Device } & \rot{W-Rider }\\
 \hline
 CenterPoint$\dagger$~\cite{yin2021center} & 13.5& 61.0& 36.0& 33.0& 28.0& 26.0& 25.0& 22.5& 16.0& 16.0& 12.5& 9.5& 8.5& 7.5& 8.0& 8.0& 7.0& 6.5& 3.0& 2.0& 14.0& 14.0& 1.0& 0.5& 0& 3.0& 0 \\
 
 CenterPoint$\ddagger$ & 22.0& 67.6& 38.9& 46.5& 16.9& 37.4& 40.1& 32.2& 28.6& 27.4& 33.4& 24.5& 8.7& 25.8& 22.6& 29.5& 22.4& 6.3& 3.9& 0.5& 20.1& 22.1& 0& 3.9& 0.5& 10.9& 4.2 \\
 
 %FSD & 28.2& 68.1& 40.9& 59.0& 29.0& 38.5& 41.8& 42.6& 39.7& 26.2& 49.0& 38.6& 20.4& 30.5& 14.8& 41.2& 26.9& 11.9& 5.9& 13.8& 33.4& 21.1& 0& 9.5& 7.1& 14.0& 9.2 \\

 VoxelNeXt~\cite{chen2023voxelnext} & 30.7& 72.7& 38.8& 63.2& 40.2& 40.1& 53.9& 64.9& 44.7& 39.4& 42.4& 40.6& 20.1& 25.2& 19.9& 44.9& 20.9& 14.9& 6.8& 15.7& 32.4& 16.9& 0& 14.4& 0.1& 17.4& 6.6\\
 
 FSF~\cite{li2023fully} & 33.2& 70.8& 44.1& 60.8& 27.7& 40.2& 41.1& 50.9& 48.9& 28.3& 60.9& 47.6& 22.7& 36.1& 26.7& 51.7& 28.1& 12.2& 6.8& 25.0& 41.6& -& -& -& -& -& - \\
 \hline
 FSD~\cite{fan2022fully} & 28.2  & 68.1  & 40.9  & 59.0  & 29.0  & 38.5  & 41.8  & 42.6  & 39.7  & 26.2  & 49.0  & 38.6  & 20.4  & 30.5  & 14.8  & 41.2  & 26.9  & 11.9  & 5.9  & 13.8  & 33.4  & 21.1  & 0  & 9.5  & 7.1  & 14.0  & 9.2 \\

 % FSD* & 29.8& 71.5& 40.5& 66.4& 31.2& 37.2& 51.5& 52.7& 43.2& 25.8& 51.5& 41.6& 18.8& 30.4& 19.9& 45.5& 27.5& 12.1& 5.6& 15.1& 35.4& 20.0& 0& 2.1& 1.9& 17.4& 9.1 \\
 
 FSD$^{*}$+RE+MVP & 31.9& 74.5& 42.1& 69.2& 32.0& 39.1& 53.3& 58.7& 50.1& 27.4& 55.9& 46.1& 21.2& 30.5& 20.2& 54.8& 26.4& 12.0& 6.2& 15.9& 42.6& 21.4& 0& 1.9& 2.3& 17.2& 9.5 \\
\hline
 % below is FSDv2 trained by authors
 FSDv2~\cite{fan2023fsdv2} & 37.6& 77.0& 47.6& 70.5& \textbf{43.6}& \textbf{41.5}& 53.9& 58.5& 56.8& 39.0& 60.7& 49.4& 28.4& \textbf{41.9}& 30.2& 44.9& \textbf{33.4}& \textbf{16.6}& \textbf{7.3}& \textbf{32.5}& 45.9& \textbf{24.0}& \textbf{1.0}& 12.6& \textbf{17.1}& \textbf{26.3}& \textbf{17.2} \\
 % % below is FSDv2 trained by us
 % FSDv2* & 37.6& 77.2& 46.9& 71.0& \textbf{43.3}& 41.4& 55.0& 60.7& 58.7& 46.4& 61.6& 50.5& 30.8& 39.9& 28.2& 47.0& 30.9& \textbf{16.5}& 7.1& \textbf{33.5}& 46.4& 23.0& 0.0& \textbf{15.7}& 4.3& \textbf{26.6}& \textbf{15.9} \\
 
 FSDv2$^{*}$+RE+MVP & \textbf{38.5}& \textbf{78.5}& \textbf{48.4}& \textbf{72.8}& 43.0& 41.5& \textbf{56.1}& \textbf{65.1}& \textbf{60.2}& \textbf{48.8}& \textbf{63.8}& \textbf{53.2}& \textbf{29.6}& 40.6& \textbf{30.3}& \textbf{55.9}& 31.3& 16.2& 7.3& 30.3& \textbf{48.9}& 23.8& 0& \textbf{14.6}& 1.4& 25.0& 14.7 \\
 \hline 
 \hline 
\end{tabular} 
\label{table:sota_comparison}
\end{table*}

% The comparison of different state-of-the art algorithms is shown in \Cref{table:sota_comparison}. Our algorithms, marked with $^{*}$, outperform their respective baselines, as well as all previous state-of-the-art methods.
We compare our best performing model (Baseline$^*$+RE+\gls{mvp}) to various state-of-the art algorithms from the literature, as shown in \Cref{table:sota_comparison}. We use FSD and FSDv2 as our baselines. The evaluation range is set to 0-200~\si{m} to enable a fair comparison to other methods.
Notably, our models outperform their respective baselines and surpass all previously proposed state-of-the-art methods.
%algorithms. Furthermore, RangeFSDv2 outperforms all previous \gls{lidar} based state-of-the-arts. 
Although improvements span across many classes, they are especially notable for 
%Notable improvements are observed for 
vulnerable road users (2-3\% for pedestrians, motorcyclist, bicyclist) and small objects (3-10\% for motorcycle, bicycle and construction cones). Equally interesting is the drop in performance for some classes, as compared to \gls{fsd}v2 baseline. 
%The reason for this gap can be attributed to the image segmentor generating the virtual points, specifically the dataset it's trained on,  As mentioned
This drop could be caused by the noise in the virtual point generation process, which in turn is caused by noisy masks generated by the 2D segmentor, for classes it is not trained on. 
%This drop can be attributed to the image segmentor generating the virtual points. 
As mentioned earlier, we use an off-the-shelf 2D segmentor (HTC~\cite{chen2019hybrid}) trained on the NuImages dataset. 
% which has 10 classes. 
We perform a one-to-one mapping between NuImage classes and their best match among the 26 classes in the \gls{av2} dataset. Inevitably, some classes not covered by this scheme don't improve (like box truck), or even get worse if mislabeled (eg. wheelchair, sign, stop sign etc.). 
%On the other hand, even though there is no explicit label for motorcyclists and bicyclists in NuImages, the segmentor probably detects them as pedestrians and correctly generates the virtual points. 
%%This underscores our design choice of omitting the semantic label from the virtual point feature, compared to original implementation of MVP~\cite{yin2021multimodal}. 

% \input{tables/Custom_Table_AP} % Not sure if this should go here

%\input{tables/Custom_Table_AR} % Not sure if this should go here

% \input{tables/Custom_Table_all_AR} % Not sure if this should go here

% \input{tables/Custom_Table_Experiments_AP}

\subsection{Ablation Studies}   \label{subsec:ablation_studies}

\textbf{Runtime Comparison:}
\begin{table}[h] 
\centering 
\scriptsize
% \footnotesize 
\caption{Runtime (\si{msec}) and memory (GB) comparison. $*$: re-trained by us.} 
\label{tab:runtime}
\renewcommand{\arraystretch}{1.2}
\setlength{\tabcolsep}{3pt}\begin{tabular}{l|cc}
 \hline 
  Method & Runtime (\si{msec}) & Memory (GB)  \\
 \hline
 FSD$^*$ &  72 & 2.9 \\
 FSD* + RE &  164 & 3.5 \\ 
 Mask RCNN + FSD* + MVP & 505 & 14.5 \\ 
 HTC + FSD* + MVP & 1800 & 27.9 \\ 
 \hline 
\end{tabular} 
\end{table}

The runtime is evaluated on a single A-100 GPU. Following~\cite{fan2023fsdv2}, we set the batch size to 1 and evaluate the average runtime for 1000 samples on the ~\gls{av2} validation set, without runtime optimizations.
The range experts incur more than twice the runtime cost, and comparable memory consumption compared to FSD. This considers a naive implementation where the mid-range and long-range experts are run sequentially. 
% Their memory consumption though is comparable to FSD
On the other hand, one could run the mid-range and long-range experts in parallel, in which case their runtime will be comparable to FSD and the memory consumption will double.
% On the other hand, \gls{mvp} requires masks from an image segmentation network for generating virtual points, which can take from 425~\si{msec} and 1720~\si{msec} respectively for  

\gls{mvp} relies on masks generated by an image segmentation network for virtual point generation. On the \gls{av2} dataset with 7 cameras, this process takes approximately 425~\si{msec} for Mask RCNN and 1720~\si{msec} for HTC networks, respectively. 
While it is important to note that this analysis may depend on specific hardware and implementation factors, the key insight remains clear: when hardware availability is not a constraint, MVP emerges as an attractive choice for improving long-range 3D detection performance. However, in cases where hardware resources are limited, range experts offer a cost-effective alternative to achieve comparable improvements.

\textbf{Effect of image segmentation:}
To analyze the impact of the image segmentation backbone on the performance of virtual point generation for \gls{mvp}, 
we replace HTC~\cite{chen2019hybrid}, the default 2D segmentor, with Mask RCNN~\cite{he2017mask}, another widely recognized image segmentation model.
The results, as displayed in \Cref{table:effect_2d_backbone}, reveal a slight improvement in 3D \gls{map} when using Mask RCNN compared to HTC, despite the latter achieving higher 2D mask and bounding box \gls{map} scores.
This observation suggests that while both segmentors exhibit robust performance in detecting mid-range objects, HTC excels in identifying smaller objects. Some of these small objects may exist at considerable distances, beyond the reach of LiDAR points, making it impossible for a \gls{lidar}-only detector to detect them. 

\begin{table}[h] 
\centering 
\scriptsize
% \footnotesize 
\caption{Effect of 2D backbone on mAP performance for FSD$^*$+MVP.} 

 \renewcommand{\arraystretch}{1.2}
\setlength{\tabcolsep}{3pt}\begin{tabular}{l|cc|c}  % \setlength{\tabcolsep}{2pt}\begin{tabular}{l|cccccccc}
 \hline 
  \multirow{2}{*}{Method}  & \multicolumn{2}{c|}{2D mAP} & \multirow{2}{*}{3D mAP}  \\
  & Mask & Box &\\
 \hline
 HTC~\cite{chen2019hybrid} & 46.4 & 57.3 & 31.5 \\ \hline
 Mask RCNN~\cite{he2017mask} & 38.4 & 47.8 & 31.8 \\ 
 \hline 
\end{tabular} 
\label{table:effect_2d_backbone}
\end{table} 

%  \renewcommand{\arraystretch}{1.2}
% % \setlength{\tabcolsep}{3pt}\begin{tabular}{l|c|c|c|c|c|c|c|c} 
% \setlength{\tabcolsep}{3pt}\begin{tabular}{l|cc|c|ccccc}  % \setlength{\tabcolsep}{2pt}\begin{tabular}{l|cccccccc}
%  \hline 
%   \multirow{2}{*}{Method}  & \multicolumn{2}{c|}{2D mAP} & \multicolumn{6}{c}{3D mAP} \\
%   \cline{4-9}
%   & Mask & Box & 0-250 & 0-50 & 50-100 & 100-150 & 150-200 & 200-250 \\
%  \hline
%  HTC~\cite{chen2019hybrid} & 46.4 & 57.3 & 31.9 & 46.9 & 16.9 & 5.8 & 2.0 & 0.5\\ \hline
%  Mask RCNN~\cite{he2017mask} & 38.4 & 47.8 & 31.8 & 46.8 & 16.9 & 5.6 & 1.9 & 0.5\\ 
%  \hline 
% \end{tabular} 
% \end{table} 

% \input{sections/EXPERIMENTS/strategy_for_training_long_range}

% \input{sections/EXPERIMENTS/performance_breakdown}

% \input{sections/EXPERIMENTS/comparison_to_baseline}

% \input{sections/EXPERIMENTS/effect_2d_backbone}

%%%%%%%%%%%%%%%%%%%%%%%%%%%%%

\section{Limitations} \label{sec:limitations}
% % Three notable limitations of our proposed method merit discussion. 
% Firstly, we did not account for extrinsic calibration mismatches between \gls{lidar} and cameras. 
% % The possibility of mismatches in extrinsic calibration between \gls{lidar} and cameras must be accounted for. 
% Such misalignments can significantly affect object detection, particularly during depth completion, leading to erroneous depth estimations or even the exclusion of objects due to the absence of nearby lidar points. This limitation becomes especially pronounced for distant objects and can have a significant impact on our network's performance. 
Firstly, the effects of fusing data from active \gls{lidar} sensors with passive camera sensors, most of which employ rolling shutters, need to be further investigated. This limitation is especially pertinent when dealing with highly dynamic objects, as they can appear differently in each sensor's data, leading to further mismatch during depth completion.  
Also, we did not investigate scenarios where large objects, like trucks with trailers, span our range boundaries. In such cases, short-range and long-range networks may yield separate bounding boxes for the same object, necessitating fusion. 
%While this was not observed in the \gls{av2} dataset, it is a crucial aspect for future development to ensure system robustness and completeness.

%Consequently, discrepancies in sensor data fusion may arise, influencing the accuracy of object detection and tracking. Addressing these limitations in future research will enable more robust and accurate multi modal object detection methodologies.
%%%%%%%%%%%%%%%%%%%%%%%%%%%%%

\section{Conclusions and Future Work} 
\label{sec:conclusions}
% \textcolor{red}{Write conclusions here}
In this work, we identify two key problems that 3D Object Detectors face at long range: \gls{lidar} sparsity, causing a form of domain gap, and label imbalance between mid and long range objects. To address these problems, 
% we propose a combination of fully sparse detectors (FSD and FSDv2) as range experts, trained using an adaptive range weighing scheme, and virtual points obtained from MVP, a powerful method for image based depth completion. Our network achieves a state-of-the-art performance on the \gls{av2} dataset. 
we analyze two solutions: a combination of fully sparse detectors (FSD and FSDv2) as range experts, trained using an adaptive range weighing scheme, and augmenting virtual points to \gls{lidar} data obtained from MVP, a powerful method for image based depth completion.
Nevertheless, a large margin for improvement remains for ranges beyond 50 meters. As a future work, our analysis could be extended to other depth completion methods, as well as cover various 3D object detectors. 
Another possible next step could be investigating whether deep image features could aid sparse \gls{lidar} features for detection, while being computationally efficient. An interesting line of investigation can be studying camera based methods to detect objects beyond the range of current \gls{lidar} sensors. 

% By expanding the perceptual horizon, we can unlock the full potential of self-driving cars, empowering them to navigate highways with enhanced situational awareness, make informed decisions, and ultimately contribute to safer and more efficient transportation systems. 
%%%%%%%%%%%%%%%%%%%%%%%%%%%%%


%%%%%%%%%%%%%%%%%%%%%%%%%%%%%%%%%%%%%%%%%%%%%%%%%%%%%%%%%%%%%%%%%%%%%%%%%%%%%%%%

%\FloatBarrier
%%%%%%%%%%%%%%%%%%%%%%%%%%%%
\section*{ACKNOWLEDGMENT}
\label{sec:acknowledgment}
%%%%%%%%%%%%%%%%%%%%%%%%%%%%
% Our special thanks to Magnus Granström for supporting us during the work.
The computations were enabled by the supercomputing resource Berzelius provided by National Supercomputer Centre at Linköping University and the Knut and Alice Wallenberg Foundation, Sweden.

%\addtolength{\textheight}{-12cm}
\bibliographystyle{IEEEtran}
\bibliography{mybib}

\end{document}